\documentclass[conference]{IEEEtran}
\usepackage{cite}
\usepackage{amsmath,amssymb,amsfonts}
\usepackage{algorithmic}
\usepackage{graphicx}
\usepackage{textcomp}
\usepackage{booktabs}   
\usepackage{lipsum}     
\usepackage{pifont}
\usepackage{xcolor}
\usepackage{multirow} 

\usepackage[most]{tcolorbox}
\usepackage{tabularx}
\usepackage{colortbl}      
\usepackage[table,dvipsnames]{xcolor}
\usepackage{tikz}
\usepackage{pgfplots}
\pgfplotsset{compat=1.18}
\usepackage{caption} 

\def\BibTeX{{\rm B\kern-.05em{\sc i\kern-.025em b}\kern-.08em
    T\kern-.1667em\lower.7ex\hbox{E}\kern-.125emX}}
\begin{document}

\title{HarmoniAD: Harmonizing Local Structures and Global Semantics for Anomaly Detection}

\author{
Naiqi Zhang$^{1}$$^{\S}$\quad
Chuancheng Shi$^{2}$$^{\S}$\quad
Jingtong Dou$^{2}$\quad
Wenhua Wu$^{2}$\quad
Fei Shen$^{3}$\quad
Jianhua Cao$^{1}$$^\ast$\\
$^{1}$Tianjin University of Science and Technology \quad
$^{2}$The University of Sydney \quad
$^{3}$National University of Singapore\\
$^{\S}$ Equal contribution \\
$^\ast$ Corresponding author: caojh@tust.edu.cn

}

\maketitle

\begin{abstract}
Anomaly detection is crucial in industrial product quality inspection. Failing to detect tiny defects often leads to serious consequences. Existing methods face a structure-semantics trade-off: structure-oriented models (such as frequency-based filters) are noise-sensitive, while semantics-oriented models (such as CLIP-based encoders) often miss fine details. To address this, we propose HarmoniAD, a frequency-guided dual-branch framework. Features are first extracted by the CLIP image encoder, then transformed into the frequency domain, and finally decoupled into high- and low-frequency paths via an adaptive cutoff for complementary modeling of structure and semantics. The high-frequency branch is equipped with a fine-grained structural attention module (FSAM) to enhance textures and edges for detecting small anomalies, while the low-frequency branch uses a global structural context module (GSCM) to capture long-range dependencies and preserve semantic consistency. Together, these branches balance fine detail and global semantics. HarmoniAD further adopts a multi-class joint training strategy, and experiments on MVTec-AD, VisA, and BTAD show state-of-the-art performance with both sensitivity and robustness.
\end{abstract}

\begin{IEEEkeywords}
Anomaly Detection, Frequency-Guided Learning, Structural Attention, Semantic Consistency
\end{IEEEkeywords}

\section{Introduction}
\begin{figure}[t]
    \centering
    \includegraphics[width=0.9\linewidth]{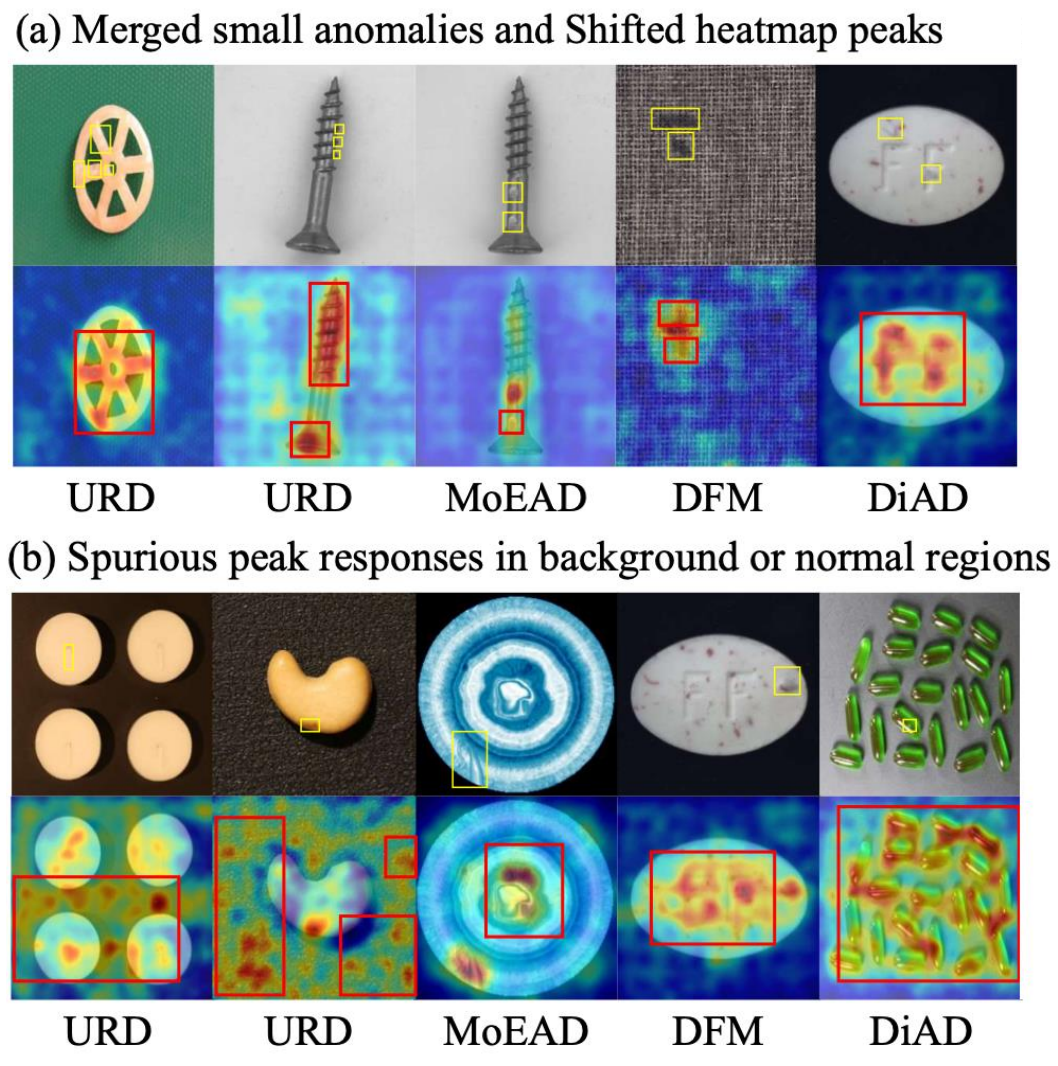}
    \caption{Examples of failure cases in existing anomaly detection methods. Yellow boxes denote true anomalies, and red boxes indicate false positives.}
    \label{fig:motivation}
    \vspace{-0.4cm}
\end{figure}

Anomaly detection \cite{chen2024implicit, park2024neural, bhattacharya2025towards} is a fundamental task in computer vision, with broad applications in industrial inspection, medical imaging, and various safety-critical systems. However, existing methods suffer from an imbalance between structure and semantics: structure-oriented models tend to be overly sensitive to noise \cite{10222596}, while semantics-oriented models often fail to detect subtle defects \cite{ma2025aa, li2025kanoclip}. As shown in Fig~\ref{fig:motivation}, systematic analysis of heatmaps reveals two recurring problems: spurious activations frequently occur in normal background regions, resulting in false alarms, and adjacent micro-defects are often wrongly merged into ambiguous areas, causing the response center to deviate from the actual anomaly. These issues reflect fundamental limitations in modeling structural boundaries, which restrict the effective distinction between anomalies and background as well as among different anomalies in complex scenarios. Therefore, harmonizing structural sensitivity with semantic consistency is key to advancing the capability of anomaly detection.

Existing anomaly detection approaches can be broadly categorized into two groups: methods that emphasize local structural sensitivity and those that rely on high-level semantic representations. The first group focuses on capturing fine-grained cues such as textures, edges, and patch-level irregularities. Typical instantiations include contrastive patch representation learning or continuous memory-based modeling. And enhanced VAE variants. While these techniques successfully highlight local anomalies, they often overfit to noise and lack global context, leading to false alarms in complex scenes.

The second group of methods builds upon high-level semantic representations, often leveraging large-scale vision-language models such as CLIP~\cite{radford2021learning}. These approaches introduce anomaly awareness through prompt engineering or semantic alignment. For instance, AA-CLIP~\cite{ma2025aa} enhances zero-shot anomaly detection by constructing anomaly-aware textual anchors to refine cross-modal alignment. Similarly, KanoCLIP~\cite{li2025kanoclip} incorporates knowledge-driven prompt learning and enhanced cross-modal integration to discriminate anomalies better. While these methods demonstrate impressive generalization and semantic reasoning ability, they tend to overlook subtle structural irregularities such as small scratches or fine-grained texture deviations.

\begin{figure*}[t]
    \centering
    \includegraphics[width=0.95\linewidth]{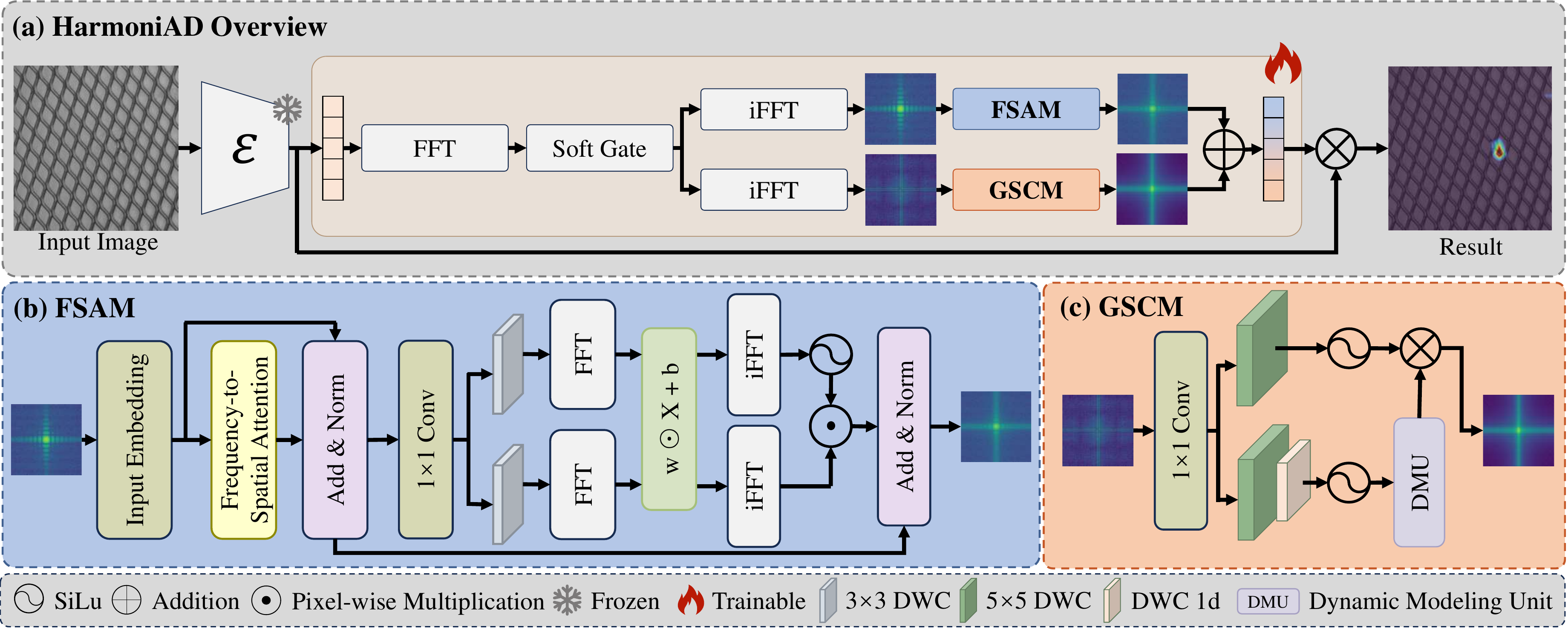}
    \caption{Overall framework of HarmoniAD. CLIP image embeddings are first transformed into the frequency domain and split by a soft gate into high- and low-frequency streams. These streams are reconstructed by the fine-grained structural attention module (FSAM) and the global structural context module (GSCM), respectively, and then perceived as the final representation. The perceived representation is contrasted with the original embeddings to derive patch-level anomaly scores.}
    \label{fig:framework}
    \vspace{-0.4cm}
\end{figure*}

To address these limitations, we propose HarmoniAD, a frequency-guided dual-branch framework that fuses local structural details with global semantic context. A frequency-domain partition separates high- and low-frequency components, enabling structural textures and semantic dependencies to be modeled in parallel. The two branches cooperate to balance fine-grained localization and semantic coherence, producing discriminative and robust anomaly representations. This structural-semantic synergy yields heatmaps that are both precise and consistent, and achieves state-of-the-art performance on multiple anomaly detection benchmarks.
\begin{itemize}
  \item We propose HarmoniAD, a frequency-guided dual-branch framework that unifies fine-grained sensitivity and global semantic consistency via complementary high- and low-frequency paths.
  \item We design a fine-grained structural attention module (FSAM), utilizing frequency-spatial interactions to enhance texture and edge sensitivity for the detection of subtle anomalies.
  \item We develop a global structural context module (GSCM) that dynamically models semantic dependencies to ensure global consistency.
\end{itemize}

\section{Related Work}

\subsection{Traditional Anomaly Detection}
Image anomaly detection (IAD) targets identifying samples or regions that deviate from the normal data distribution under unsupervised or weakly supervised settings. Prior work mainly follows two paradigms: representation or distribution modeling methods, such as PaDiM~\cite{defard2021padim}, SPADE~\cite{yoon2022spade}, and PatchCore~\cite{roth2022towards}, which model normality in the feature space of pretrained vision encoders via distribution estimation or memory banks and localize anomalies using distributional discrepancies or nearest-neighbor distances; and generative or reconstruction methods, including AE and VAE variants and GAN-based methods, which treat reconstruction residuals as anomaly cues. Complementary directions include synthetic anomaly augmentation with self-supervised learning, exemplified by CutPaste~\cite{li2021cutpaste}, and diffusion-based denoising priors integrated into reconstruction and localization pipelines, such as DiAD~\cite{he2024diffusion}. Despite strong performance in many industrial scenarios, these approaches often rely on local appearance deviations and can be insufficient for anomalies that require global semantic understanding and contextual coherence, motivating the adoption of high-level semantic priors such as CLIP~\cite{radford2021learning}.

\subsection{CLIP-based Anomaly Detection}
CLIP is a large-scale vision-language model that has been pre-trained using image-text alignment. It provides transferable semantic representations for anomaly detection. Representative methods such as WinCLIP~\cite{jeong2023winclip} and PromptAD~\cite{li2024promptad} perform label-free anomaly detection and localization by introducing class-specific textual prompts and matching them to image features via similarity scoring; however, their performance is sensitive to prompt design and phrasing. More importantly, while CLIP-based approaches benefit from semantic alignment for improved generalization, they typically do not explicitly disentangle or jointly model the local structural cues and the global semantic dependencies embedded across multiple layers of the visual encoder. This often yields a pronounced trade-off between detecting subtle, fine-grained structural anomalies and maintaining semantic consistency.

\section{Methods}

\subsection{Overview} 
We propose HarmoniAD, a frequency-guided dual-branch framework that jointly models fine-grained structures and global semantics (Fig~\ref{fig:framework}a). A frozen CLIP encoder extracts high-level features, which are projected into the frequency domain; a differentiable Soft Gate adaptively separates high- and low-frequency components according to their frequency radius for end-to-end, structure-aware routing. Unlike multi-scale or multi-level decoupling that primarily varies spatial resolution or receptive fields while keeping structure and semantics entangled in the same representation, our frequency-domain split explicitly separates structural details and global semantic components within a single embedding for more controllable specialization. For local anomalies, the fine-grained structural attention module (FSAM) (Fig~\ref{fig:framework}b) enhances texture and edge sensitivity via frequency-spatial attention, while the global structural context module (GSCM) (Fig~\ref{fig:framework}c) with a DMU captures semantic dependencies to maintain global consistency.

\subsection{Adaptive High- and Low-Frequency Separation via Soft Gate}
\label{sec:softgate}
To avoid the non-differentiability of hard threshold selection, we treat the cutoff as a latent scale variable over the discrete candidate set $\{r_m\}_{m=1}^M$ and define a Gibbs distribution
$p_m=\mathrm{Softmax}_m\!\big(\kappa J(r_m)\big)$.
We then obtain an input-adaptive boundary as its expectation,
\begin{equation}
c=\sum_{m=1}^M r_m\,p_m
=\sum_{m=1}^M r_m\,\mathrm{Softmax}_m\!\big(\kappa J(r_m)\big).
\end{equation}
Here $\kappa>0$ acts as an inverse temperature, trading off between near-point selection and distributional averaging; consequently, scale partitioning becomes data-conditioned inference rather than a fixed design choice.

\subsection{Fine-grained Structural Attention Module}
The fine-grained structural attention module (FSAM) enhances local anomaly detection by leveraging frequency-domain priors for structure-aware feature refinement. Unlike generic attention that merely reweights features within a single spatial representation, FSAM injects frequency-derived structural priors via frequency-to-spatial attention (F2S Attn) and amplitude modulation, explicitly strengthening boundary and texture cues for pixel-level localization.

\noindent\textbf{Frequency-to-Spatial Attention (F2S Attn)}\;
We add a scalar relative bias to the attention logits via a 4D offset descriptor
$\Delta\mathbf p_{ij}=[x_i-x_j,\;y_i-y_j,\;|x_i-x_j|,\;|y_i-y_j|]$ with $\mathbf p_i=(x_i,y_i)$.
Let $\mathbf e_\theta\in\mathbb R^{4}$, $\beta_\theta(\Delta\mathbf p_{ij})=\mathbf e_\theta^\top\Delta\mathbf p_{ij}$, and $\mathbf B=[\beta_\theta(\Delta\mathbf p_{ij})]_{i,j}\in\mathbb R^{N_f\times N_s}$:
\begin{equation}
\operatorname{Attn}(\mathbf Q,\mathbf K,\mathbf V)
=\operatorname{Softmax}\!\Bigl(\tfrac{\mathbf Q\mathbf K^\top}{\sqrt d}+\mathbf B\Bigr)\mathbf V.
\end{equation}
Here $\mathbf Q\in\mathbb R^{N_f\times d}$ and $\mathbf K,\mathbf V\in\mathbb R^{N_s\times d}$; $[\cdot]_{i,j}$ indexes all $(i,j)$ pairs.

\noindent\textbf{Frequency-domain Channel Modulation.}
Given $X$ be the result of the Fourier transform, with amplitude $A=|X|$ and phase $\Phi=\arg(X)$. A nonnegative mask $m \in \mathbb{R}_{+}^{C\times H\times W}$ modulates the amplitude:
\begin{equation}
\widehat{A} = m \odot \sigma(A), 
\quad 
\widehat{X} = \widehat{A} \odot \tfrac{X}{A+\varepsilon},
\end{equation}
where $\sigma(\cdot)$ is a nonlinearity, $\odot$ denotes Hadamard product, and $\varepsilon > 0$ ensures stability. The output is obtained via inverse Fourier transform of $\widehat{X}$, preserving phase and adaptively enhancing structural details through amplitude modulation.

\subsection{Global Structural Context Module}
We design a lightweight module to capture long-range dependencies with dynamic modulation and coordinate encoding. In contrast to standard non-local or Transformer context blocks with static aggregation, GSCM performs token-wise dynamic modulation and uses a DMU to couple low-frequency semantics with the upper stream, suppressing spurious activations while enforcing global semantic coherence.
Given $X=[x_1,\dots,x_T] \in \mathbb{R}^{T\times C}$,$S \in \mathbb{R}^{T\times T}$ and $B^{\mathrm{rel}} \in \mathbb{R}^{T\times T}$, 
the module first computes a dynamic affinity:
\begin{equation}
S = \mathrm{Softmax}\!\Bigl(\tfrac{XW_q (XW_k)^\top}{\sqrt{r}} + B^{\mathrm{rel}}\Bigr),
\end{equation}
where $W_q,W_k\in\mathbb{R}^{C\times r}$ are learnable, $r<C$, 
and $B^{\mathrm{rel}}=\{b_{ij}^{\mathrm{rel}}\}$ encodes relative positional bias.
Each token $x_t$ generates dynamic weights $\Pi_t=\mathrm{diag}(\xi(W_\pi x_t))$ and bias $\Gamma_t=W_\gamma x_t$, 
with $W_\pi,W_\gamma\in\mathbb{R}^{C\times C}$ and nonnegative activation $\xi(\cdot)$.  
The aggregated representation is then
\begin{equation}
\begin{aligned}
Z = \mathcal{A}\!\left(S\,[(\Pi_1x_1+\Gamma_1),\dots,(\Pi_Tx_T+\Gamma_T)]\right),
\end{aligned}
\end{equation}
where $\mathcal{A}(\cdot)$ is a nonlinearity and $Z \in \mathbb{R}^{T\times C}$.  
We introduce a dynamic modeling unit (DMU) to endow the lower branch with cross-token dynamics and to multiplicatively couple it with the upper branch:
\begin{equation}
x_t = W_o\!\big(u_t \odot \operatorname{SiLU}(v_t + m_t)\big).
\end{equation}
The modulation term is gated:
\begin{equation}
m_t = g_t \odot s_{t-1} + (1-g_t)\odot \phi\!\big(W_d[\,r_t\odot v_t\,]\big).
\end{equation}
Here $u_t$ (upper-branch conv feature), $v_t$ (lower-branch feature) and $s_{t-1}$ are in $\mathbb{R}^C$; $g_t=\sigma(W_g^{\mathrm{d}} v_t)$, $r_t=\sigma(W_r v_t)$; $W_{\cdot}$ learnable, $\sigma,\phi$ pointwise, $\odot$ is Hadamard. Finally, a gate $\mathrm{Gate}=\sigma(XW_g^{\mathrm{o}})\in\mathbb{R}^{T\times C}$ 
and coordinate encoding $\mathrm{Coord}=[W_c\rho(p_1),\dots,W_c\rho(p_T)]$ 
are fused with $Z$ to produce
\begin{equation}
Y = \mathrm{Gate}\odot Z + \mathrm{Coord}. 
\end{equation}

\subsection{Reconstruction and Loss Supervision}
\noindent\textbf{Reconstruction.}
Let $\hat{F}_{\mathrm{high}}, \hat{F}_{\mathrm{low}} \in \mathbb{R}^{C\times H\times W}$ denote the two branch outputs (Sec.~\ref{sec:softgate}). We fuse them with $\mathcal{P}_h$  and  $\mathcal{P}_l$ in $[0,1]$:
\begin{equation}
\label{eq:fusion}
X_{\mathrm{recon}}
= \mathcal{P}_h(\hat{F}_{\mathrm{high}})
+ \mathcal{P}_l(\hat{F}_{\mathrm{low}}).
\end{equation}
When $\mathcal{P}_h=\mathcal{P}_l=\mathcal{I}$  (the identity mapping), the fusion degenerates to a weighted summation.

\noindent\textbf{Loss supervision.}  
The overall objective consists of six complementary loss terms: the cosine reconstruction losses for normal images and regions (\(\mathcal{L}^{cos}_{n}\), \(\mathcal{L}^{cos}_{an}\)), the cosine reconstruction loss and push-away loss for abnormal regions (\(\mathcal{L}^{cos}_{a}\), \(\mathcal{L}_{far}\)), as well as the spatially-aware contrastive loss (\(\mathcal{L}_{con}\)) and triplet loss (\(\mathcal{L}_{tri}\)). Here, \(\theta\) denotes the model parameters and \(\lambda\) represents the weighting coefficients for each loss term. These components collectively enhance the separation, clustering, and localization of normal and abnormal features.
Formally,
\begin{equation}
\begin{aligned}
\mathcal{L}_{\mathrm{total}} =
&\ \lambda_{n}\,\mathcal{L}_{n}^{\cos}
+ \lambda_{a}\,\mathcal{L}_{a}^{\cos}
+ \lambda_{con}\,\mathcal{L}_{con}
+ \lambda_{\mathrm{an}}\,\mathcal{L}_{\mathrm{an}}^{\cos} \\
&+ \lambda_{\mathrm{far}}\,\mathcal{L}_{\mathrm{far}}
+ \lambda_{\mathrm{tri}}\,\mathcal{L}_{\mathrm{tri}}
+ \lambda_{\mathrm{reg}}\,\|\theta\|_2^2.
\end{aligned}
\end{equation}

\begin{table*}[t]
\centering
\caption{\textbf{Comparison of seven methods for unified anomaly detection on three datasets.} Each dataset is evaluated by four indicators: P-ROC, I-ROC, P-PR, and I-PR. All metrics are the higher the better.}
\label{tab:comparison_methods_as_rows}
\vspace{-0.2cm}
\renewcommand{\arraystretch}{1.0}
\begin{tabular*}{\linewidth}{@{\extracolsep{\fill}}c|cccccccccccc}
\hline
\hline
& \multicolumn{4}{c}{\textbf{MVTec-AD}~\cite{bergmann2019mvtec}}
& \multicolumn{4}{c}{\textbf{VisA}~\cite{zou2022spot}}
& \multicolumn{4}{c}{\textbf{BTAD}~\cite{btad2020}} \\
\textbf{Method}& P-ROC & I-ROC & P-PR & I-PR
& P-ROC & I-ROC & P-PR & I-PR
& P-ROC & I-ROC & P-PR & I-PR \\
\hline
MoEAD~\cite{meng2024moead}       & 97.0 & 97.7 & 43.8 & 97.9 & 98.7 & 93.1 & 34.2 & 93.7 & 97.1 & 92.3 & 51.3 & 98.2 \\
URD~\cite{liu2025unlocking}    & 95.8 & 90.8 & 47.4 & 96.7 & 97.0 & 91.5 & 33.9 & 93.7 & 98.5 & 92.4 & 59.3 & 98.1 \\
DFM~\cite{fre2023} & 96.5 & 69.7 & 42.4 & 89.8 & 96.5 & 51.6 & 25.2 & 77.8 & 96.3 & 68.8 & 48.0 & 82.8 \\
DiAD~\cite{he2024diffusion}     & 96.8 & 97.2 & 49.0 & 96.9 & 96.0 & 86.8 & 24.3 & 90.2 & 96.9 & 92.0 & 47.9 & 94.4 \\
WinCLIP~\cite{jeong2023winclip}   & 81.4 & 71.6 & 17.8 & 84.5 & 73.8 & 66.1 & 5.34 & 71.1 & 66.7 & 55.2 & 7.3 & 62.7 \\
PromptAD~\cite{li2024promptad}  & 95.4 & 91.4 & 49.3 & 95.8 & 96.7 & 85.5 & 27.8 & 87.5 & 96.5 & 90.0 & 55.5 & 94.2 \\
KanoCLIP~\cite{li2025kanoclip}  & 93.1 & 94.3 & -- & -- & 83.8 & 97.7 & -- & -- & 90.6 & 96.5 & -- & -- \\
\textbf{HarmoniAD (Ours)} & \textbf{98.0} & \textbf{98.0} & \textbf{58.3} & \textbf{99.5} & \textbf{98.9} & \textbf{94.3} & \textbf{44.2} & \textbf{95.6}& \textbf{98.9} & \textbf{94.4} & \textbf{60.9} &\textbf{98.8}  \\
\hline
\hline
\end{tabular*}
\vspace{-0.3cm}
\end{table*}

\begin{figure*}[t]
    \centering
    \includegraphics[width=0.9\linewidth]{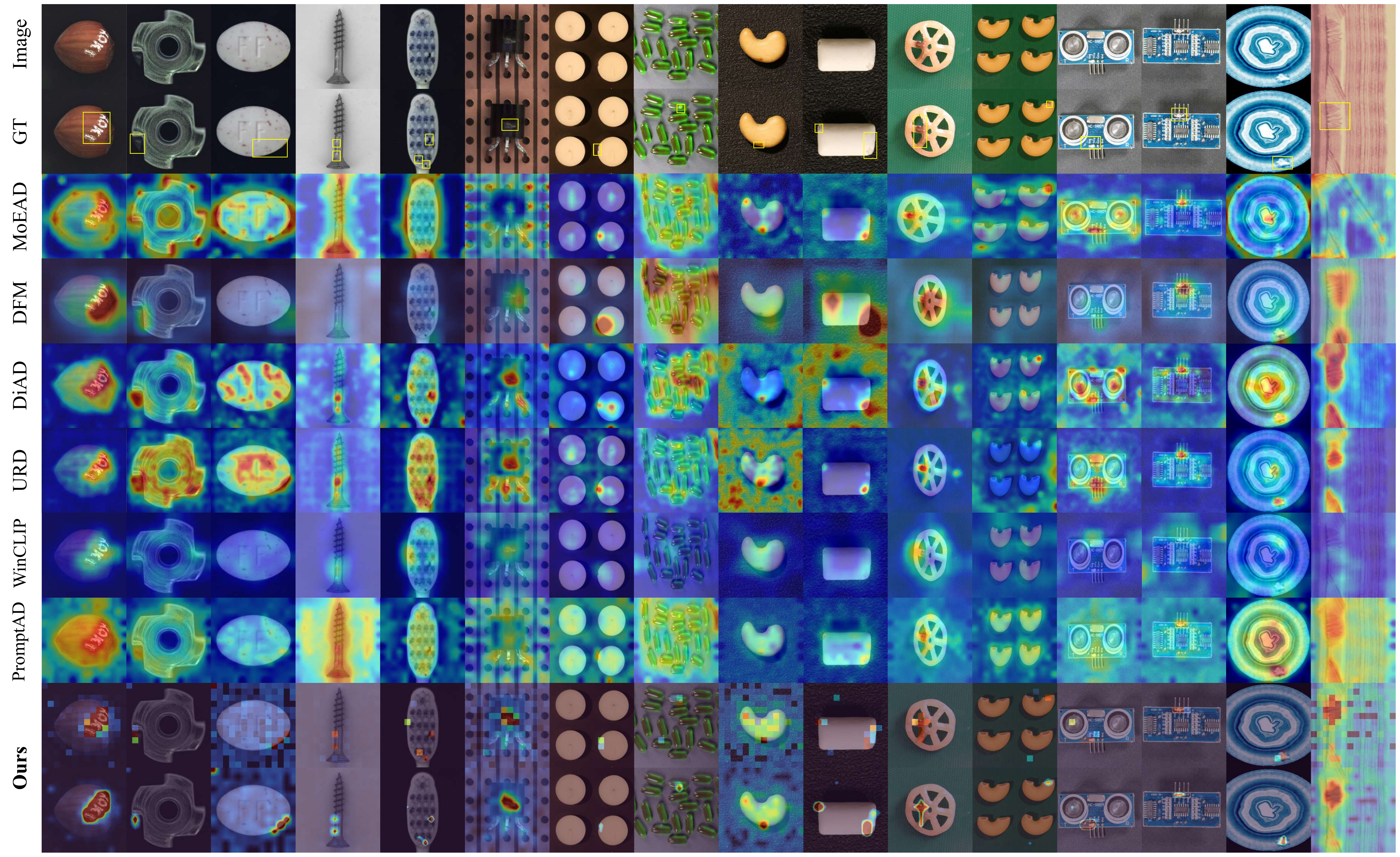}
    \caption{With multi-class joint training, anomaly localization results are presented on selected categories from the MVTec-AD, VisA, and BTAD datasets. Each column is a test sample. The first row shows ground-truth defects (yellow boxes). Rows 3-8 are heatmaps from comparison methods. Rows 9-10 show our patch-level and pixel-level heatmaps. Our method yields high responses in anomaly regions and low responses elsewhere, and outperforms comparison methods on challenging cases.}
    \label{fig:mvtec_results}
    \vspace{-0.4cm}
\end{figure*}

\section{Experiments And Analysis}

\subsection{Implementation Details}
\noindent\textbf{Datasets.} Following DiAD~\cite{he2024diffusion}, we adopt the same training pipeline and data splitting strategy for multi-class joint training, ensuring a fair and consistent comparison with prior methods. We evaluate HarmoniAD on three widely used industrial anomaly detection benchmarks: MVTec-AD~\cite{bergmann2019mvtec}, VisA~\cite{zou2022spot}, and BTAD~\cite{btad2020}.

\noindent\textbf{Metrics.} We evaluate HarmoniAD against state-of-the-art methods using ROC and PR metrics at both image and pixel levels. Specifically, I-ROC and I-PR assess image-level anomaly detection accuracy, while P-ROC and P-PR evaluate pixel-level anomaly localization performance.

\noindent\textbf{Hyperparameters.} We use ViT-B/16 as the frozen CLIP backbone with $224 \times 224$ inputs, trained with Adam (batch size 36, learning rate $1 \times 10^{-2}$) on a single NVIDIA A100-40GB GPU.

\subsection{Compare with SOTA Methods}
We quantitatively and qualitatively compare the proposed HarmoniAD with several representative SOTA methods in a multi-class joint training setting, including MoEAD~\cite{meng2024moead}, URD~\cite{liu2025unlocking}, DFM~\cite{fre2023}, DiAD~\cite{he2024diffusion}, WinCLIP~\cite{jeong2023winclip}, PromptAD~\cite{li2024promptad}, KanoCLIP~\cite{li2025kanoclip} and our method consistently achieves the best results across all benchmarks.

\noindent\textbf{Quantitative Results.} To evaluate unified anomaly detection, we conduct quantitative experiments on MVTec-AD, VisA, and BTAD. As shown in Table~\ref{tab:comparison_methods_as_rows}, our method consistently achieves the best performance across all datasets. Specifically, on MVTec-AD, our approach achieves a P-PR score of 58.3, surpassing the strongest competitor by 9.3 points, demonstrating its superior capability in suppressing false activations and accurately localizing defects. On VisA, our method improves I-ROC from 93.1 to 94.3, indicating more reliable image-level anomaly detection. Similar performance gains are observed on BTAD, where our method attains the highest scores on all four metrics. These improvements indicate that HarmoniAD better balances structural sensitivity and semantic consistency. We attribute the gains to our frequency-guided dual-stream specialization, which explicitly isolates high-frequency structural evidence for precise localization and low-frequency semantic coherence for mitigating background-induced false activations, a separation that is typically not enforced by multi-level feature decoupling or standard attention/context modules.

\noindent\textbf{Qualitative Results.} As shown in Fig~\ref{fig:mvtec_results}, we compare HarmoniAD with several SOTA methods on MVTec-AD. MoEAD often overemphasizes high-frequency edges and textures, leading to hollow or ring-like responses, while DFM produces concentrated yet frequently misaligned activations and misses subtle defects. DiAD improves sensitivity to small anomalies but suffers from widespread false positives, and URD, although suppressing background noise, tends to merge nearby defects and triggers on complex textures. In contrast, HarmoniAD yields more accurate and compact localization, with clearer boundaries and better separation of small defects. Moreover, its coarse-to-fine predictions remain visually consistent across patch- and pixel-level outputs. These improvements stem from adaptive frequency separation and dual-branch reconstruction that better exploit CLIP semantic features.

\begin{table}[t]
\centering
\caption{Soft Gate Ablation Study on BTAD}
\label{tab:freq_ablation}
\begin{tabular}{l|cccc}
\hline
\hline
\textbf{Threshold} & \textbf{P-ROC} & \textbf{I-ROC} & \textbf{P-PR} & \textbf{I-PR} \\
\hline
0.3 & 96.8 & 90.1 & 52.5 & 91.8 \\
0.5 & 98.2 & 91.7 & 54.4 & 94.0 \\
0.7 & 97.7 & 91.3 & 52.9 & 92.5 \\
\textbf{Soft Gate} & \textbf{98.9} & \textbf{94.4} & \textbf{60.9} & \textbf{98.8} \\
\hline
\hline
\end{tabular}
\vspace{-0.3cm}
\end{table}

\begin{table}[t]
\centering
\caption{\textbf{Ablation on BTAD.} Ablation of FSAM and GSCM (FSAM uses F2S Attn by default).}
\label{tab:module_ablation_only}
\renewcommand{\arraystretch}{1.0}
\begin{tabular}{cc|cccc}
\hline
\hline
\textbf{FSAM} & \textbf{GSCM} & \textbf{P-ROC} & \textbf{I-ROC} & \textbf{P-PR} & \textbf{I-PR}\\
\hline
\textcolor{gray}{\ding{55}} & \ding{51} & 97.4 & 93.4 & 53.2 & 90.5 \\
\ding{51} & \textcolor{gray}{\ding{55}} & 97.8 & 92.6 & 52.1 & 92.6 \\
\hline
\ding{51} & \ding{51} & \textbf{98.9} & \textbf{94.4} & \textbf{60.9} & \textbf{98.8} \\
\hline
\hline
\end{tabular}
\vspace{-0.5cm}
\end{table}

\subsection{Ablation Study}

\noindent\textbf{Ablation of Soft Gate.} To validate the effectiveness of our adaptive high- and low-frequency splitting module (Soft Gate), we conduct an ablation study on BTAD by comparing it against hard splits with fixed thresholds $t\in\{0.3, 0.5, 0.7\}$. As shown in Table~\ref{tab:freq_ablation}, fixed thresholds exhibit noticeable performance sensitivity to $t$, whereas Soft Gate achieves the best results across all four metrics and remains consistently superior even to the strongest fixed setting, with particularly larger gains in PR and AUROC. These results indicate that adaptive soft partitioning more robustly allocates high- and low-frequency contributions, thereby improving anomaly detection performance.

\noindent\textbf{Ablation of FSAM and GSCM.} To validate the effectiveness and complementarity of the fine-grained structural attention module (FSAM) and the global structural context module (GSCM), we conduct a module-level ablation study on the BTAD dataset (Table~\ref{tab:module_ablation_only}). The results highlight the complementary contributions of local structural modeling and global semantic constraints. With Soft Gate enabled, removing FSAM while retaining GSCM preserves image-level discrimination (I-ROC = 93.4) but substantially degrades pixel-level localization (P-PR = 53.2), indicating reduced sensitivity to fine-grained anomalies. Conversely, retaining FSAM while removing GSCM improves local detection (P-ROC = 97.8) at the cost of global consistency (I-ROC = 92.6), suggesting increased background interference. When both modules are enabled, the model achieves the best overall performance, with P-ROC, I-ROC, P-PR, and I-PR reaching 98.9, 94.4, 60.9, and 98.8, respectively. These results demonstrate that FSAM enhances fine-grained structural sensitivity. At the same time, GSCM suppresses spurious activations and enforces global coherence, and their joint integration yields a balanced and robust anomaly detection framework.

\begin{table}[t]
  \centering
  \caption{\textbf{Ablation Study on BTAD.} Effect of enabling F2S Attn while keeping FSAM and GSCM active.}
  \label{tab:attn_ablation_only}
  \renewcommand{\arraystretch}{1.3}
  \begin{tabular}{c|cccc}
    \hline
    \hline
    \textbf{F2S Attn.} & \textbf{P-ROC} & \textbf{I-ROC} & \textbf{P-PR} & \textbf{I-PR}\\
    \hline
    \textcolor{gray}{\ding{55}}  & 98.0 & 93.5 & 56.6 & 95.7 \\
    \ding{51}  & \textbf{98.9} & \textbf{94.4} & \textbf{60.9} & \textbf{98.8} \\
    \hline
    \hline
  \end{tabular}
  \vspace{-0.5cm}
\end{table}

\begin{figure}[t]
    \centering
    \includegraphics[width=0.85\linewidth]{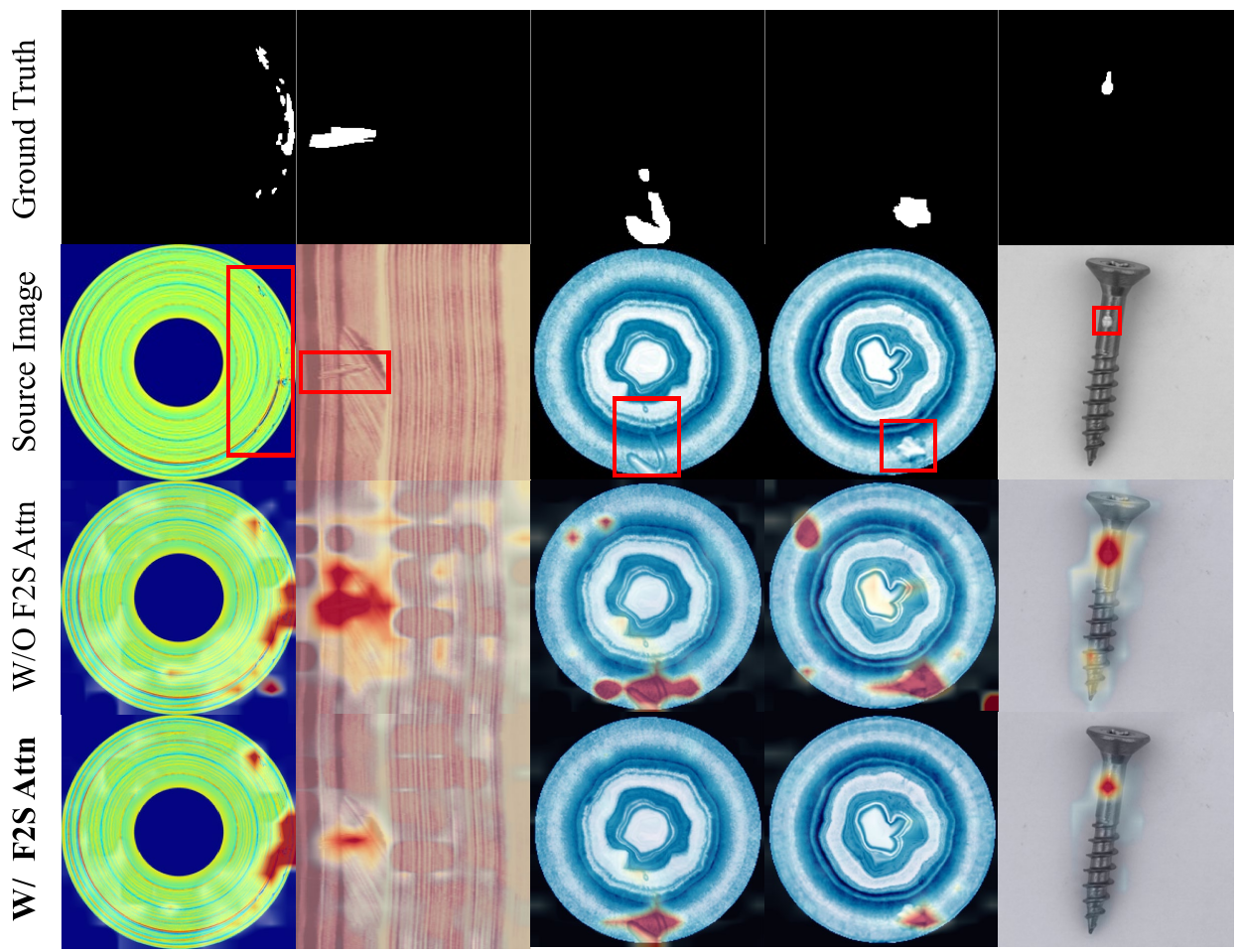}
    \caption{Ablation study of F2S Attn. Red boxes denote true anomalies.}
    \label{fig:abl_attn}
    \vspace{-0.6cm}
\end{figure}

\begin{figure}[t]
    \centering
    \includegraphics[width=0.9\linewidth]{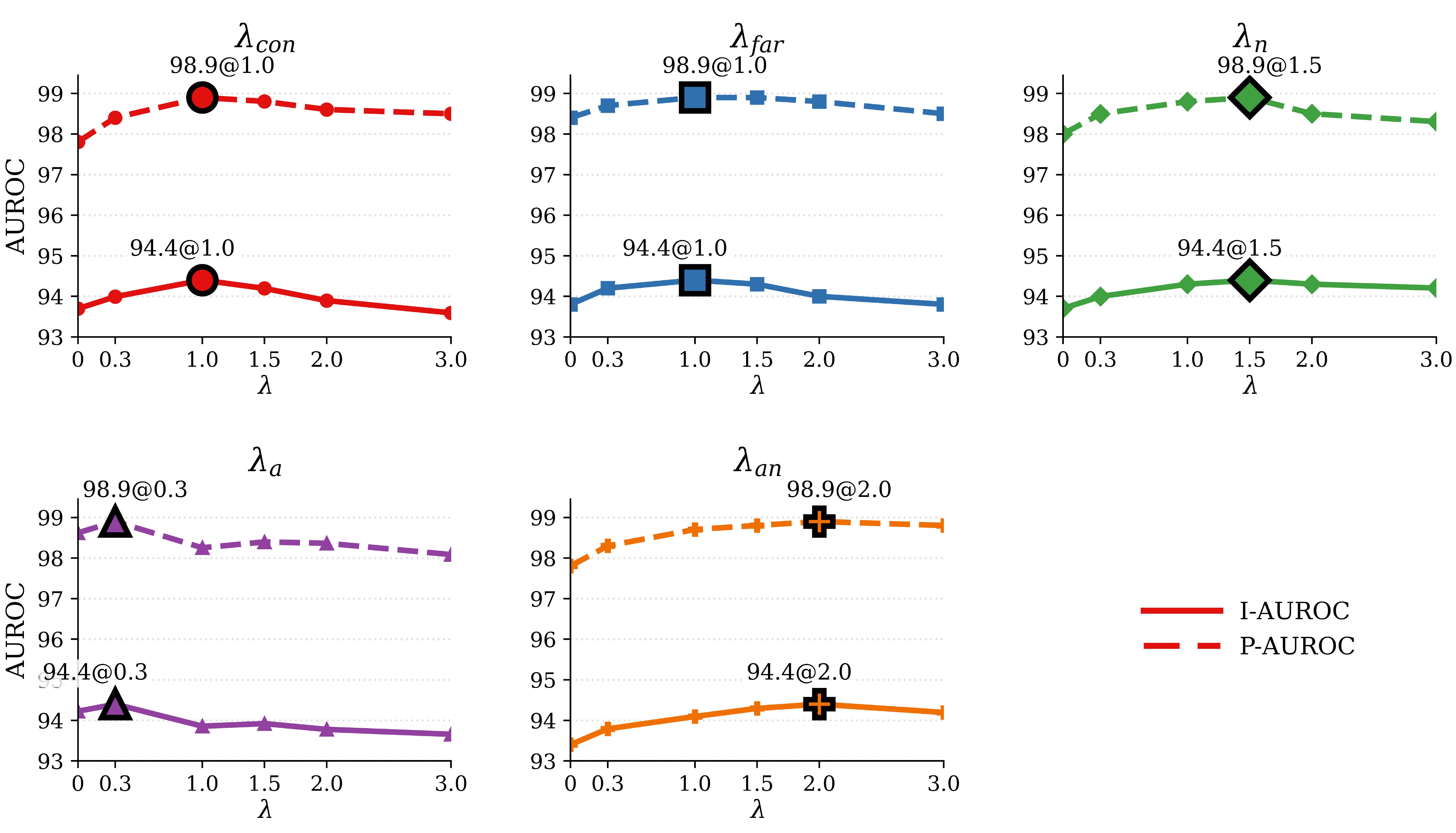}
    \caption{Sensitivity analysis of core loss weights on BTAD.}
    \label{fig:lambda}
    \vspace{-0.5cm}
\end{figure}

\noindent\textbf{Ablation of F2S Attn.} We further evaluate the contribution of frequency-to-spatial attention (F2S Attn) through an ablation study on the BTAD dataset with FSAM and GSCM enabled. As shown in Table~\ref{tab:attn_ablation_only} and Fig~\ref{fig:abl_attn}, disabling F2S Attn causes consistent performance degradation across all metrics (P-ROC 98.0, I-ROC 93.5, P-PR 56.6, I-PR 95.7) and results in spatially diffuse, noisy anomaly responses with ambiguous boundaries. In contrast, enabling F2S Attn yields clear improvements, particularly at the pixel level, with P-PR and I-PR increasing by 4.3 and 3.1 percentage points, respectively, and produces more compact, concentrated activation maps that align well with ground-truth defects. These quantitative and qualitative results demonstrate that F2S Attn effectively leverages frequency-domain cues to guide spatial attention, enhancing fine-grained anomaly localization while maintaining global consistency.

\subsection{Parameter Sensitivity Analysis}
To assess the robustness of our method with respect to the weights of different loss terms, we conduct a hyperparameter sensitivity analysis on the BTAD dataset by varying each loss weight while keeping the others fixed. Fig~\ref{fig:lambda} reports the performance trends in terms of image-level and pixel-level AUROC. The results show that performance remains stable across a wide range of loss weights, indicating that the proposed framework does not require careful hyperparameter tuning. For each loss term, performance exhibits a clear but smooth peak around the selected default value, whereas deviations from this optimum lead to only marginal performance degradation. In particular, both image-level and pixel-level AUROC curves demonstrate consistent trends, suggesting that the loss components contribute in a complementary and well-balanced manner.

\section{Conclusion}
In conclusion, we presented HarmoniAD, a frequency-guided dual-branch framework for anomaly detection. Through adaptive frequency decoupling, it jointly captures local details and global semantics, enabling accurate and efficient anomaly localization. Extensive benchmarks confirm its state-of-the-art performance, validating frequency-domain structural modeling. These results also indicate that frequency-domain structural cues provide a principled and interpretable signal for anomaly characterization. Future work will extend HarmoniAD with temporal modeling for video anomaly detection.

\bibliographystyle{IEEEbib}
\bibliography{ref}
\newpage
\newpage
\section*{Supplementary Material}

\appendix

The appendices provide additional details that support and extend the main paper. 
Appendix~\ref{more_exp} presents further experimental results and ablation studies.
Appendix~\ref{diss} addresses common issues.
Appendix~\ref{lim} covers the limitations of our work.

\section{More Experiments}
\label{more_exp}

\begin{figure*}[t]
    \centering
    \includegraphics[width=0.95\linewidth]{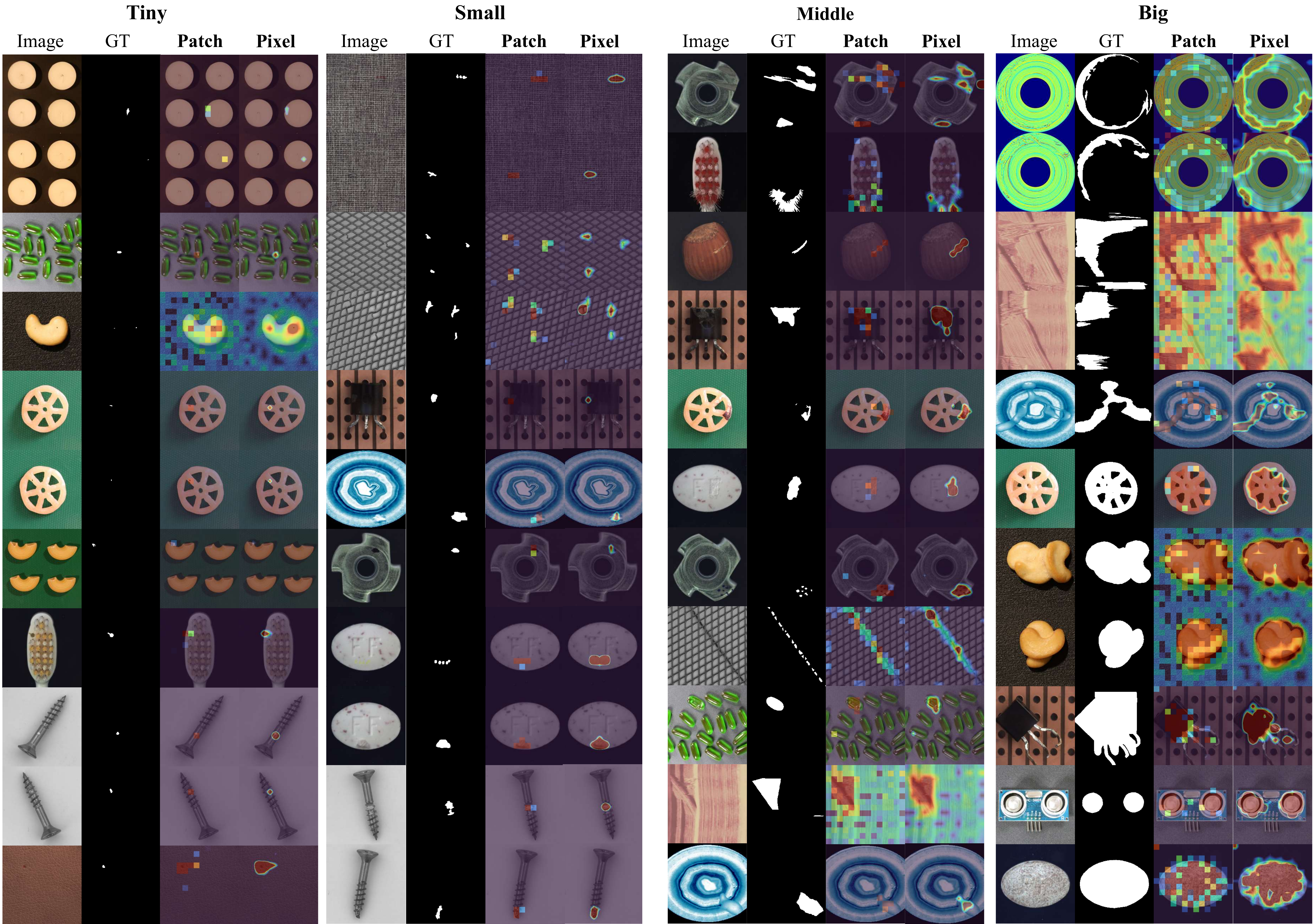}
    \caption{Heatmaps across anomaly area scales. Columns are grouped by anomaly area scale (Tiny, Small, Middle, Big). Within each group, the column order is Image / GT / Patch / Pixel. Patch denotes the patch-level output heatmap (upsampled to the image space), while Pixel denotes the pixel-level output heatmap. Heatmap colors indicate anomaly scores using the same color mapping as in the main text.}
    \label{fig:more}
\end{figure*}

\subsection{Qualitative Results Under Different Anomaly Area Scales} 

This appendix presents additional qualitative visualizations at different anomaly-area scales (Tiny / Small / Middle / Big). For each scale group, we provide the input image (Image), the ground-truth mask (GT), and the predicted heatmaps at two output granularities (Patch and Pixel). To ensure fair comparison, all visualizations follow the same pipeline as in the main paper (including normalization, upsampling, and consistent color mapping).

\subsection{Cross-domain Evaluation} 
\begin{figure*}[!htbp]
    \centering
    \includegraphics[width=0.9\linewidth]{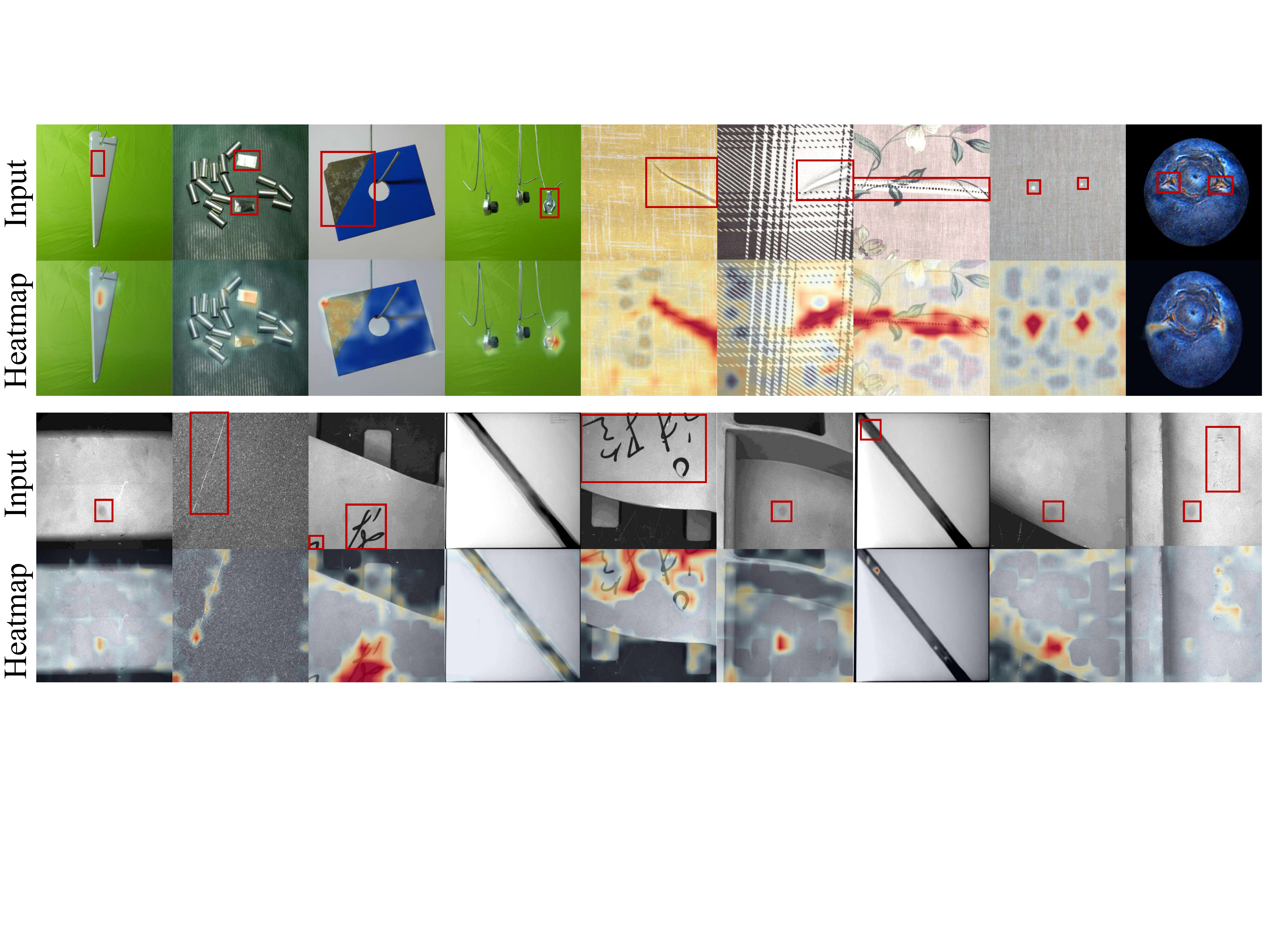}
    \caption{Cross-domain zero-shot inference and localization. Anomalous regions are marked with red bounding boxes. For each sample, the top image shows the input image and the bottom image shows the zero-shot inference heatmap.}
    \label{fig:cross}
    \vspace{-0.4cm}
\end{figure*}

We further evaluate cross-domain generalization by performing zero-shot inference with a single set of weights jointly trained on BTAD, MVTec-AD, and VisA. Without any dataset-specific adaptation, we apply the trained model to out-of-domain benchmarks (MPDD and WFDD) as well as anomaly samples we collected from real industrial production, and additionally include a new semantic category from agriculture (blueberry defects). Figure~\ref{fig:cross} presents representative cross-domain heatmaps: the first row shows results on MPDD and WFDD, with the rightmost example corresponding to blueberry defects, while the second row consists entirely of our collected real-world industrial anomalous devices/components. Despite clear distribution shifts, previously unseen categories, and abnormal patterns not observed during training, the model often produces plausible anomaly responses, where higher activations concentrate around suspected defective regions and provide reasonably informative localization cues. Notably, in the blueberry cases, the model can still highlight abnormal surface areas to some extent, suggesting non-trivial transferability beyond the industrial domains seen during training. Overall, these qualitative results indicate a certain degree of cross-domain robustness in a pure zero-shot setting, while a more comprehensive quantitative analysis is left for future work.

\subsection{Inference Latency}
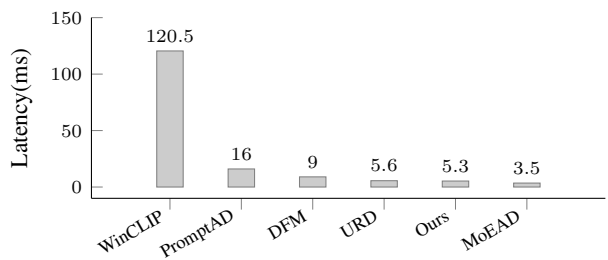
\begin{figure}[!htbp]
\label{fps}
  \centering
  \begin{tikzpicture}
    \begin{axis}[
      ybar,
      width=0.95\linewidth,
      height=0.45\linewidth,
      bar width=10pt,
      ymin=-10, ymax=150,                 
      ytick={0,50,100,150,190},
      enlarge x limits=0.22,              
      axis x line*=bottom,
      axis y line*=left,
      ylabel={Latency(ms)},
      symbolic x coords={WinCLIP,PromptAD,DFM,URD,Ours,MoEAD}, 
      xtick=data,
      xticklabel style={rotate=30, anchor=east},
      nodes near coords,
      every node near coord/.append style={
        font=\scriptsize,
        /pgf/number format/fixed,
        /pgf/number format/precision=1
      },
      tick label style={font=\scriptsize},
      label style={font=\small},
    ]
      \addplot[fill=black!20, draw=black!50] coordinates {
        (WinCLIP,120.5)
        (PromptAD,16.0)
        (DFM,9.0)
        (URD,5.6)
        (Ours,5.3)
        (MoEAD,3.5)
      };
    \end{axis}
  \end{tikzpicture}
  \caption{Latency comparison.}
  \label{fig:fps_compare}
\end{figure}

Figure~\ref{fig:fps_compare} reports the inference throughput in Latency. Our method achieves throughput comparable to the lightweight high-efficiency baselines, while being substantially faster than other CLIP-based methods in this comparison, indicating that it maintains strong efficiency without introducing a noticeable speed overhead.

\section{Discussions}
\label{diss}
\noindent$\triangleright$ \textbf{\textit{Q1. Why do we report PR and AUROC rather than AUPRO?}} \\
AUROC and PR are threshold-agnostic, ranking-based metrics that can be computed under a unified protocol at both the image and pixel levels. AUROC measures the overall separability between normal and anomalous score distributions, while PR is more sensitive to false positives under severe anomaly sparsity and better captures the precision–recall trade-off.  In contrast, AUPRO requires thresholding anomaly maps into connected regions and integrating performance over a specified FPR range. It is highly sensitive to the threshold, connected-component definitions, and post-processing such as smoothing, as well as region morphology, which can entangle evaluation design choices with method contributions and understate improvements on small-scale and boundary anomalies.

\noindent$\triangleright$ \textbf{\textit{Q2. Why is the ablation study conducted only on BTAD rather than on others?}} \\
We benchmark HarmoniAD on all three datasets, but we conduct systematic ablations only on BTAD because it is more diagnostic for isolating the effects of individual design choices. MVTec AD and VisA are larger benchmarks with diverse categories and both image- and pixel-level annotations. However, recent industrial anomaly detection results on these datasets are often near ceiling, which compresses the observable gaps caused by toggling components and makes attribution less reliable. In our setting, BTAD provides larger headroom and clearer sensitivity to architectural changes, yielding higher signal-to-noise component-level validation.

\noindent$\triangleright$ \textbf{\textit{Q3. Why do we freeze the CLIP backbone rather than fine-tuning it, and how sensitive is HarmoniAD to the backbone choice?}} \\
We treat the backbone as a stable general-purpose feature extractor and attribute the primary performance gains to our frequency decomposition and the proposed structure semantics coordination modules. Freezing CLIP ViT serves two purposes. This substantially lowers training cost and mitigates overfitting risks that often arise when fine-tuning on limited data for certain categories. We additionally tested other ViT-based pretrained backbones, such as DINOv2, and observed only minor metric differences, suggesting that HarmoniAD is not highly sensitive to the specific backbone. Under a joint consideration of accuracy, inference speed, and memory footprint, frozen CLIP provides the best overall cost effectiveness and is therefore used as the default configuration.

\section{Limitation} 
\label{lim}
The proposed method is designed for single-frame image representations and does not introduce explicit temporal modeling. We therefore do not conduct video anomaly detection experiments, where temporal dependencies and consistency are often essential for handling evolving anomalies, normal motion, and dynamic backgrounds. Future work will extend our frequency based structure semantics division to the spatiotemporal setting by incorporating temporal relation modeling and temporal consistency regularization, aiming for a dedicated video anomaly detection framework and evaluation.

\end{document}